\documentclass[pmlr]{jmlr}

\RequirePackage{graphicx}
 \usepackage{booktabs}
 \usepackage{natbib}  
\usepackage{caption} 
\usepackage{algorithm}
\usepackage{algorithmic}
\usepackage{enumitem}
\usepackage{multirow}
\usepackage{amsmath}
\usepackage{amssymb}
\usepackage{booktabs}
\usepackage{newfloat}
\usepackage{listings}

\newcommand{\Prob}{\mathbb{P}}

\usepackage{longtable}
\makeatletter
\def\set@curr@file#1{\def\@curr@file{#1}} 
\makeatother
\usepackage[load-configurations=version-1]{siunitx} 

\theorembodyfont{\upshape}
\theoremheaderfont{\scshape}
\theorempostheader{:}
\theoremsep{\newline}

\makeatletter
\def\@titlefoot{}
\def\ps@jmlrtps{%
  \let\@mkboth\@gobbletwo
  \def\@oddhead{}%
  \let\@evenhead\@oddhead
  \def\@oddfoot{}%
  \let\@evenfoot\@oddfoot
}
\makeatother

\title[Self-Harm Risk Screening via Adaptive Multi-Agent LLM Systems]{Reliable Self-Harm Risk Screening via Adaptive Multi-Agent LLM Systems}

\author{\Name{Meghana Karnam} \and \Name{Ananya Joshi}\\
\addr Johns Hopkins University\\ Baltimore, USA}

\begin{document}

\maketitle

\begin{abstract}
 Emerging AI systems in behavioral health and psychiatry use multi-step or multi-agent LLM pipelines for tasks like assessing self-harm risk  and screening for depression. However, common evaluation approaches, like LLM-as-a-judge, do not indicate when a decision is reliable or how errors may accumulate across multiple LLM judgements, limiting their suitability for safety-critical settings. We present a statistical framework for multi-agent pipelines structured as directed acyclic graphs (DAGs) that provides an alternative to heuristic voting with principled, adaptive decision-making. We model each agent as a stochastic categorical decision and introduce (1) tighter agent-level performance confidence bounds, (2) a bandit-based adaptive sampling strategy based on input difficulty, and (3) regret guarantees over the multi-agent system that shows logarithmic error growth when deployed. We evaluate our system on two labeled datasets in behavioral health : the AEGIS 2.0 behavioral health subset (N=161) and a stratified sample of SWMH Reddit posts (N=250). Empirically, our adaptive sampling strategy achieves the lowest false positive rate of any condition across both datasets, 0.095 on AEGIS 2.0 compared to 0.159 for single-agent models, reducing incorrect flagging of safe content by 40\% and still having similar false negative rates across all conditions. These results suggest that principled adaptive sampling offers a meaningful improvement in precision without reducing recall in this setting.
\end{abstract}

\section{Introduction}

Hospitals and behavioral health organizations are exploring AI systems to support psychiatric intake, triage, and crisis response \citep{weber2026can, pavlopoulos2024overview, sun2025practical}. In high-risk settings such as self-harm screening, these systems must follow strict, institution-specific standard operating procedures (SOPs) that define how risk is assessed, when escalation is required, and what constitutes a safe response. Missing self-harm risk in a patient interaction can result in a crisis response being delayed or missed entirely, with potentially life-threatening consequences. Still, excessive escalation of safe content to human review creates alert fatigue and strains already limited clinical resources~\citep{ancker2017effects}. Recent work proposes operationalizing these SOPs using multi-agent LLM systems, where each specialized agent handles a distinct clinical responsibility (e.g., risk detection, safety planning, escalation) and uncertain cases are passed along structured 
workflows~\citep{care-ad}. Modular agent designs mirror real-world care coordination and improve robustness over single models, making them especially relevant to behavioral health, where decision-making is inherently staged, heterogeneous, and safety-critical. In practice, a resident or volunteer may screen patient-facing messages first, uncertain cases proceed to a supervisor for secondary review, and edge cases reach a compliance officer before any content reaches 
a patient.

Still, current multi-agent systems are largely heuristic. In practice, they rely on simple majority voting over a fixed number of stochastic LLM calls (LLM-as-a-judge or LLM-as-a-jury) for each decision. This raises a fundamental challenge: there is no principled way to know when a decision is reliable, how many samples are sufficient, or how uncertainty propagates through the pipeline. In addition, in behavioral health settings, errors are asymmetric. A false negative (failing to detect self-harm risk) can result in unsafe responses reaching a patient in crisis, while excessive escalation increases clinician burden and system cost. Designing these systems therefore requires calibrated trade-offs between accuracy, safety, latency, and resource use. Without formal guarantees, practitioners must rely on ad hoc choices (e.g., “sample 5 times”), which may cause any of these failure modes. As a result, existing empirical evaluations are insufficient for deployment, auditing, or policy design because they do not provide input-dependent guarantees or worst-case bounds -- they only provide point estimates of performance.

This motivates the need for a statistical empirical evaluation framework that addresses:\\ (A)\textit{When should a behavioral health AI system act on its own assessment of patient risk, and when should it defer to a human clinician for review?}\\
(B) \textit{How can a system reduce missed detections of self-harm risk without increasing the burden on clinicians through unnecessary escalations?}\\
(C) \textit{How do system-level errors accumulate over time as the number of patient interactions grows under real-world deployment?}

\paragraph{Contributions.}
\begin{itemize}[leftmargin=1.5em]

    \item \textbf{Reliable decision-making at each step of care:}
    We introduce a method that quantifies how confident the multi-agent system is in each risk-sensitive decision, enabling it to distinguish between cases it can safely resolve and those that require expert or clinician review. This provides explicit reliability guarantees at each stage of the workflow rather than relying on fixed heuristics.

    \item \textbf{Adaptive sampling based on case complexity:}
    We develop a strategy that allocates more computational effort to ambiguous or high-risk cases while resolving straightforward cases efficiently. This reduces missed high-risk cases while avoiding unnecessary escalation and clinician burden.

    \item \textbf{System-level safety under deployment:}
    We show that the overall system becomes more reliable over time, with errors growing slowly even as the number of patient interactions increases. This provides a foundation for auditing and deploying multi-agent AI systems in safety-critical behavioral health settings.

\end{itemize}

\subsection{Generalizable Insights about Machine Learning in the Context of Healthcare}
This work provides practical design principles for building reliable multi-agent AI systems in behavioral health and other clinical settings where decisions are staged and safety-critical.

First, we provide \textit{a unified representation of decision steps:} Representing multiple functions in a clinical decision in a standard format (e.g., risk levels, escalation choices) allows us to assess system-level reliability consistently across the entire workflow. This simplifies auditing and ensures that safety criteria are applied uniformly.

Second, our framework provides \textit{confidence-guided decision thresholds:} The framework links how difficult a case is (e.g., how ambiguous the patient self-harm intent is) to how much evidence is needed before taking action. This provides a concrete way to decide when the system can act autonomously versus when it should defer or escalate, replacing fixed heuristics with case-dependent thresholds.

The contributions listed above also apply broadly to clinical AI systems that involve multi-stage triage, sequential decision-making, or escalation protocols, including psychiatric intake, risk screening, and care coordination workflows.

\section{Related Work}
\label{sec:related}

Detecting suicide risk and self-harm in clinical and social media text has been studied extensively and predates LLMs. Early approaches applied machine learning to clinical notes~\citep{{huang2014detecting}, {guo2024text}}, and datasets such as ScAN~\citep{rawat-etal-2022-scan} provided expert annotations of suicide attempts and ideation from hospital records. Benchmarks including Reddit SuicideWatch and Mental Health Collection (SWMH)~\citep{ji2021mentalbert} and the CLPsych shared tasks~\citep{zirikly2019clpsych} extended this work to social media, and Columbia Suicide Severity Rating Scale (C-SSRS) grounded datasets~\citep{gaur2019knowledge} introduced 
clinically validated severity scales into NLP evaluation pipelines for suicide risk. Work on depression and self-harm detection in online forums~\citep{yates2017depression} showed that patient-generated text carries meaningful clinical signal even outside formal clinical settings. Our work uses these same datasets and clinical frameworks but asks not just whether a system gets the right answer \textit{on average}, but whether it can tell us when a specific decision 
is reliable.

Large language models have also introduced new settings for evaluating self-harm and safety-sensitive decisions in healthcare and health-adjacent text. AEGIS 2.0~\citep{ghosh-etal-2025-aegis2} provides human-annotated LLM responses across safety categories 
including self-harm based on the Suicide Ideation Detection in Social Media Forums datasets. ToxicChat~\citep{lin2023toxicchat} and RealToxicityPrompts~\citep{gehman2020realtoxicityprompts} evaluate toxicity in real user interactions and AI-generated text, and broader analyses of LLM harms~\citep{weidinger2021harms} provide a framework for understanding where these systems fail. What most of these evaluations share is a reliance on single-pass classification (llm-as-a-judge) or fixed-sample voting (llm-as-a-jury), with no way to determine how many samples are enough or how errors compound when multiple agents are involved. The most closely related system from \cite{joshi2026constrainedprocessmapsmultiagent} organizes behavioral health evaluation agents into a directed acyclic graph (DAG) and shows up to 19\% accuracy gains over single-agent approaches using Monte Carlo sampling at each node. Other multi-agent systems~\citep{yang2025agentnet} similarly route inputs 
through structured agent pipelines. Yet, none of these systems provide 
formal guarantees on when a routing decision is trustworthy, how many 
samples are sufficient, or how errors accumulate over time. 


\section{Methods}
\label{sec:methods}

\begin{figure}[t]
\centering
\includegraphics[width=0.7\columnwidth]{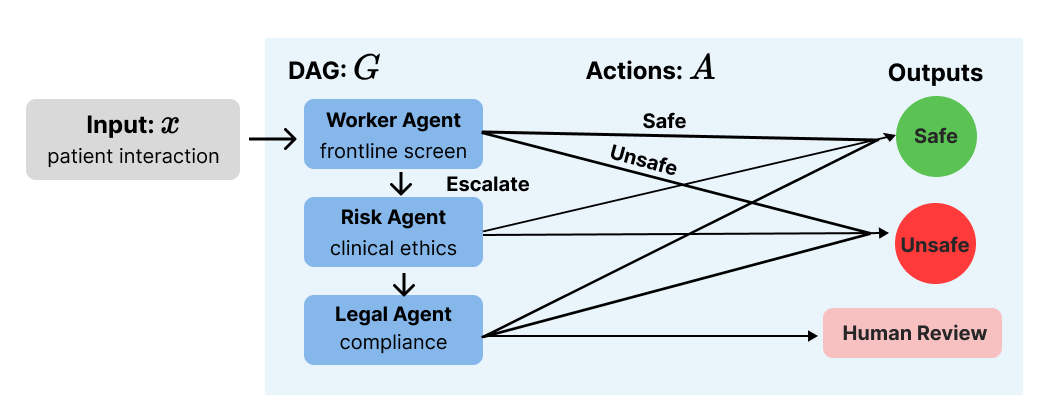}
\caption{DAG architecture of the multi-agent evaluation pipeline.Inputs that cannot be confidently resolved at the Worker node are escalated to the Risk node, then to the Legal node. If all three nodes return \texttt{escalate}, the input is referred to a human clinician for review; otherwise, a \texttt{safe} or \texttt{unsafe} label is committed at whichever node first reaches a confident decision.}
\label{fig:dag}
\end{figure}

As shown in Fig. \ref{fig:dag}, we organize LLM-based behavioral health evaluation agents into a DAG that mirrors the staged human review process used in behavioral health standard operating procedures. The DAG contains three specialized agent nodes arranged in a fixed escalation chain. The \textbf{Worker} node performs first-pass content screening, analogous to a frontline crisis counselor or resident who handles the majority of incoming content. Inputs that cannot be confidently resolved are escalated to the \textbf{Risk} node, which performs secondary review analogous to a clinical supervisor or risk assessment specialist. Cases that remain unresolved proceed to the \textbf{Legal} node, which performs final institutional compliance review before any input reaches a human clinician/expert.

\begin{table*}[h]
\centering
\small
\setlength{\tabcolsep}{4pt}

\begin{tabular}{
p{2cm} p{5cm} 
@{\hspace{.5cm}} |   
p{2cm} p{5cm}
}
\toprule
\textbf{Symbol} & \textbf{Definition} 
& \textbf{Symbol} & \textbf{Definition (contd.)} \\
\midrule

$G = (\mathcal{S}, E)$ & DAG of agent nodes \& escalation edges 
& $p_s(c\,;\,x)$ & True probability node $s$ returns label $c$ on input $x$ \\[2pt]

$\mathcal{S}$ & Set of agent nodes; $|\mathcal{S}| = 3$
& $\hat{p}_s(c\,;\,x)$ & Empirical estimate of $p_s(c\,;\,x)$ from samples \\[2pt]

$\mathcal{A}$ & Shared action space; $K = |\mathcal{A}| = 3$
& $c^*(s,x)$ & The label (\texttt{safe}, \texttt{unsafe}, or 
\texttt{escalate}) that node $s$ would most often assign to 
input $x$ if sampled many times \\[2pt]

$K$ & Number of label arms: \{\texttt{safe, unsafe, escalate}\}
& $\Delta_s(x)$ & Probability gap, defined as   $p_s(c^*) - \max_{c \neq c^*} p_s(c)$ \\[2pt]

$\tau_{\max}$ & DAG diameter; maximum episode length
& $\Delta_{\min}$ & Minimum gap over all nodes and inputs \\[2pt]

$T$ & Total evaluation episodes under deployment
& $T_c$ & Number of pulls of arm $c$ within one episode \\[2pt]

$B$ & Sampling budget per node
& $w_c$ & Confidence width for arm $c$ at current round \\[2pt]

$\delta$ & Confidence parameter; failure probability $\leq \delta$
& $\mathcal{A}_{\mathrm{act}}$ & Set of active (non-eliminated) arms \\[2pt]

$n$ & Number of fixed samples in majority voting
& & \\

\bottomrule
\end{tabular}

\caption{Notation used throughout the paper. $T$ refers to total deployment episodes in the regret theorems, while $T_c$ denotes per-arm pull counts within Algorithm~\ref{alg:se}.}
\label{tab:notation}
\end{table*}

Each node receives the original input and produces a label from the shared action space $\mathcal{A} = \{\texttt{safe}, \texttt{unsafe}, \texttt{escalate}\}$, $K = 3$ (Figure~\ref{fig:dag}). A \texttt{safe} or \texttt{unsafe} label terminates routing and commits the input to a outcome. An \texttt{escalate} label passes the input to the next node in the chain. If all three nodes return \texttt{escalate}, the input is referred to a human clinician / expert for review. Notation used throughout 
the paper is summarized in Table~\ref{tab:notation}.

\subsection{Clinical Grounding of the Action Space ($\mathcal{A}$)}

All three agent nodes are based on the C-SSRS~\citep{posner2011columbia}, a validated clinical instrument for suicide risk assessment widely adopted by hospitals and crisis intervention programs. The action space definitions are:

\noindent $\bullet$ \textbf{Safe} --- content is appropriate and supportive, consistent with safe messaging guidelines from the Suicide Prevention Resource Center (SPRC) and American Foundation for Suicide Prevention (AFSP), which emphasizes empathetic responses, psychoeducation, and appropriate crisis resources.

\noindent $\bullet$ \textbf{Unsafe} --- content poses direct patient safety risk meeting C-SSRS criteria, including any expression of suicidal ideation or self-harm intent, chatbot responses that violate safe messaging guidelines, or content a clinician would likely flag for immediate intervention.

\noindent $\bullet$ \textbf{Escalate} --- the node cannot make a confident safe or unsafe determination given available evidence; the input requires review by the next specialized agent. This label is reserved for genuine uncertainty, not content sensitivity.

\subsection{Adaptive Sampling}

Prior work from ~\citep{joshi2026constrainedprocessmapsmultiagent} uses fixed-sample majority voting at each node and involves drawing $n$ independent LLM samples and returning the plurality winner. This approach discards intermediate evidence and applies the same computational budget to every input regardless of difficulty. If the first three samples all return \texttt{unsafe}, there is no early stopping mechanism -- the system draws $n - 3$ additional samples that add no information but still incur API cost and latency.

We replace fixed majority voting, as shown in Alg. \ref{alg:se},  with an adaptive sampling bandit algorithm that allocates samples adaptively based on input difficulty. Following a setup similar to ~\cite{even-dar2006action}, each node on each input defines a $K$-armed bandit problem: arm $c \in \mathcal{A}$ has unknown mean reward $p_s(c; x)$, the true probability that node $s$ returns label $c$ on input $x$, and a pull is one LLM API call. The goal is to identify the best arm $c^* = \arg\max_c p_s(c; x)$ using as few pulls as possible within a budget $B$. However, standard Adaptive Sampling~\citep{even-dar2006action} forces a label when the budget is exhausted; our variant instead escalates, preserving the clinical safety guarantee that uncertain inputs always reach a more specialized reviewer rather than receiving a potentially wrong label.

The algorithm works by repeatedly sampling each possible label and eliminating options that are clearly not the best, stopping as soon as one label is identified with sufficient confidence. Formally, it begins by initializing all $K=3$ labels as active 
arms and continues sampling until either one arm survives or the budget $B$ is exhausted. In each round, it pulls every active arm once and updates empirical estimates (lines~3--5), then computes a confidence interval for each arm that shrinks as 
more samples are collected (line~6). An arm is eliminated when even its most optimistic estimate falls below the most pessimistic estimate of the leading arm (lines~7--8), meaning no additional sampling could plausibly reverse the ranking. When one arm survives, it is returned as the routing decision (lines~9--10). If the budget is exhausted before elimination converges, the node returns \texttt{escalate} rather than forcing a potentially wrong label (lines~11--12). This identify-or-escalate termination condition is our 
contribution to the adaptive sampling framework: 
\citet{even-dar2006action} force a label when the budget is exhausted, while our variant escalates, preserving the clinical safety guarantee that uncertain inputs always reach a more 
specialized reviewer.

\begin{algorithm}[t]
\caption{Adaptive Sampling for Node $s$, Input $x$}
\label{alg:se}
\begin{algorithmic}[1]
\REQUIRE Budget $B \in \mathbb{N}$, confidence $\delta \in (0,1)$
\STATE $\mathcal{A}_{\mathrm{act}} \leftarrow \mathcal{A}$;\
       $T_c \leftarrow 0\ \forall c \in \mathcal{A}$
\WHILE{$|\mathcal{A}_{\mathrm{act}}| > 1$ \AND $\sum_c T_c < B$}
    \FOR{each $c \in \mathcal{A}_{\mathrm{act}}$}
        \STATE Draw one LLM sample; update $\hat{p}_s(c)$,
               $T_c \leftarrow T_c + 1$
    \ENDFOR
    \STATE $w_c \leftarrow \sqrt{\dfrac{\ln(4KT_c^2/\delta)}{2T_c}}$
           for each $c \in \mathcal{A}_{\mathrm{act}}$
    \STATE $c_{\max} \leftarrow \arg\max_{c} \hat{p}_s(c)$
    \STATE Remove $c$ from $\mathcal{A}_{\mathrm{act}}$ if
           $\hat{p}_s(c_{\max}) - w_{c_{\max}} > \hat{p}_s(c) + w_c$
\ENDWHILE
\IF{$|\mathcal{A}_{\mathrm{act}}| = 1$}
    \RETURN surviving arm
\ELSE
    \RETURN \texttt{escalate}
\ENDIF
\end{algorithmic}
\end{algorithm}

\subsection{Theoretical Results}
\hypertarget{sec:theory}{}

\paragraph{Node-level correctness.}
The shared action space $\mathcal{A}$ across all three nodes/agents is the key structural choice enabling a unified theoretical treatment of the full pipeline. Without it, different inequalities with different constants would apply at each node, requiring an additional union bound penalty at the system level. With it, the same guarantee applies uniformly: the algorithm never confidently commits to a wrong label. When it cannot reach a confident 
decision, it escalates rather than guessing. Formally, with probability at least $1-\delta$, Algorithm~\ref{alg:se} never eliminates 
the true best arm $c^*$: it either returns $c^*$ or returns 
\texttt{escalate}. This guarantee rests on our definition of a Good Event 
$\mathcal{G}$ as the event that all confidence intervals contain 
the true probability at every round simultaneously:
\[
\mathcal{G} = \left\{\forall c \in \mathcal{A},\ \forall m \geq 1 :
|\hat{p}_s^{(m)}(c) - p_s(c)| \leq 
\sqrt{\frac{\ln(4KT_c^2/\delta)}{2m}}\right\}.
\]
Under $\mathcal{G}$, eliminating $c^*$ would require another 
arm to have strictly higher true probability, contradicting 
uniqueness. $\mathcal{G}$ holds with probability $\geq 1-\delta$ 
by a union bound over $K$ arms and all rounds using 
$\sum_{m=1}^{\infty} 1/m^2 = \pi^2/6 < 2$ (full proof in 
\hyperlink{app:good_event}{Appendix~B}), derived from the 
DKW inequality~\citep{dvoretzky1956asymptotic, massart1990tight} 
applied to the categorical distribution over $\mathcal{A}$ 
(derivation in \hyperlink{app:dkw}{Appendix~A}).

\paragraph{Sample complexity.}
The number of LLM calls needed to reach a confident decision 
depends on how clear the input is. A post with an unambiguous 
suicide risk signal requires fewer calls than one that is 
borderline. Formally, the minimum pulls required to identify 
$c^*$ is $n^* = 2\ln(2/\delta)/\Delta_s(x)^2$, where 
$\Delta_s(x)$ is the probability gap between the best and 
second-best label. Easy inputs resolve quickly; ambiguous ones 
exhaust the budget and escalate to human review. This provides 
the theoretical basis for the budget sweep in the results: the empirical inflection between 
$B=75$ and $B=100$ reflects where $n^*$ lies for inputs in 
this dataset (\hyperlink{app:dkw}{Appendix~A}).

\paragraph{System-level reliability over time.}
As the system processes more patient interactions, its error 
rate grows slowly rather than accumulating linearly. This 
means a hospital deploying the system at the start of a year 
will not see errors pile up proportionally with volume. 
Formally, the adaptive policy achieves $O(\log T)$ cumulative 
regret over $T$ deployment episodes compared to $O(T)$ under 
fixed majority voting, meaning doubling patient volume increases 
total errors by only $\ln 2 \approx 0.69$. The proof 
decomposes regret into a learning phase, where the algorithm 
identifies $c^*$ at each node after 
$O(K\ln(KT)/\Delta_{\min}^2)$ episodes, and an exploitation 
phase with negligible errors (proof in 
\hyperlink{app:mdp}{Appendix~C}).

\section{Experimental Setup}
\label{sec:setup}

We evaluate this adaptive sampling approach on the following datasets: 

\textbf{AEGIS 2.0}~\citep{ghosh-etal-2025-aegis2} is a content safety benchmark developed by NVIDIA comprising human-annotated LLM responses across a range of safety categories. We filter to suicide and self-harm related content across all three splits (train, validation, test), yielding N=163 unique examples after deduplication: 172 safe responses and 119 unsafe responses.\footnote{Two examples were excluded due to Unicode encoding errors in the source data that prevented API calls from completing. Results are reported over the remaining N=161 examples.} Ground truth labels are taken from the \texttt{response\_label} field, following the evaluation setup of~\citep{joshi2026constrainedprocessmapsmultiagent}. This dataset serves as our primary benchmark.

\textbf{SWMH}~\citep{ji2021mentalbert} is a Reddit mental health dataset comprising posts from SuicideWatch and four general mental health communities covering depression, anxiety, bipolar disorder, and general emotional support. We use subreddit membership as a proxy for clinical ground truth: SuicideWatch posts are labeled \texttt{unsafe}, reflecting acute suicide risk content consistent with C-SSRS criteria, and posts from the remaining four communities are labeled \texttt{safe}, reflecting general mental health discussion that does not meet criteria for immediate intervention. We sample 50 posts per subreddit for a stratified evaluation set of N=250 examples (50 unsafe, 200 safe). This dataset serves as a held-out generalization test. We evaluate these datasets across ten AI systems conditions spanning single-agent, fixed-sample, and adaptive approaches.

\begin{itemize}
    \item [1] \textbf{Single-agent} makes one LLM call to the \textsc{Worker} node only, with no DAG routing. This is the simplest baseline and represents current practice in single-agent systems.
    \item [2-4] \textbf{Majority vote} (MV $n \in \{1, 3, 5\}$) routes inputs through the full Worker-Risk-Legal DAG, making $n$ independent LLM calls at each node and returning the plurality winner. MV $n=1$ is equivalent to single-pass DAG routing; MV $n=3$ and MV $n=5$ represent increasingly expensive fixed-sample baselines.
    \item [5--10] \textbf{Adaptive Sampling} (Algorithm~\ref{alg:se}) with budgets $B \in \{10, 50, 75, 100, 124, 150\}$. $B=10$ and $B=50$ test low-budget degenerate behavior. $B=75$ tests the regime near but below the theoretical minimum effective budget. $B=100$ is the primary adaptive condition: the sample complexity bound predicts roughly 90 total pulls for inputs with $\Delta_s(x) \approx 0.5$ (\hyperlink{sec:theory}{Theoretical Results}), and $B=100$ provides a small buffer above this threshold. $B=124$ was selected as an intermediate budget between the predicted thresholds for moderate-gap inputs ($\approx 90$ pulls) and harder inputs near $\Delta_{\min} \approx 0.4$ ($\approx 138$ pulls), testing whether additional budget beyond $B=100$ benefits the harder subset of inputs. $B=150$ exceeds both theoretical thresholds to test whether further headroom yields additional gain.

\end{itemize}

\begin{table}[h]
\centering
\small
\setlength{\tabcolsep}{4pt}
\begin{tabular}{p{3cm}p{6cm}p{4.7cm}}
\toprule
\textbf{Metric} & \textbf{Definition} & \textbf{Clinical Relevance} \\
\midrule
Accuracy $\uparrow$ & Fraction of non-escalated inputs correctly labeled & Overall reliability \\[2pt]
FPR $\downarrow$ & Fraction of safe inputs labeled unsafe & Unnecessary escalation burden \\[2pt]
FNR $\downarrow$ & Fraction of unsafe inputs labeled safe & Missed crisis detection \\[2pt]
Escalation rate $\downarrow$ & Fraction of inputs reaching human review & Clinician workload \\[2pt]
Avg.\ pulls  $\downarrow$ & Total LLM calls / $N$ & Compute cost \\[2pt]
95\% CI $\downarrow$  & Wilson score interval on each proportion & Statistical reliability \\[2pt]
SW FNR  $\downarrow$ & FNR on SuicideWatch posts only & Acute crisis detection (Exp.\ 2) \\
\bottomrule
\end{tabular}
\caption{Metrics reported across all experiments. FPR and FNR are computed over non-escalated examples only. SW FNR is the primary metric for the SWMH generalization experiment.}
\label{tab:metrics}
\end{table}

We report the following metrics for each condition, computed over non-escalated examples unless otherwise noted (Table~\ref{tab:metrics}). All proportions are reported with 95\% Wilson score confidence intervals~\citep{wilson1927probable}. Full implementation details including model configuration, system prompt structure, and dataset 
preprocessing steps are provided in \hyperlink{app:reproducibility}{Appendix~D}.

\section{Results}
\label{sec:results}

\begin{table}[t]
\centering
\small
\setlength{\tabcolsep}{3pt}

\begin{tabular}{lcccc}
\toprule
\multicolumn{5}{c}{\textbf{AEGIS 2.0 (N=161)}} \\
\midrule
\textbf{Condition} & \textbf{Accuracy}$\uparrow$ & \textbf{FPR}$\downarrow$ & 
\textbf{FNR} $\star$ $\downarrow$ & \textbf{Esc.} $\downarrow$ \\
\midrule
Single-agent
  & 0.752 [0.679, 0.813]
  & 0.159 [0.079, 0.294]
  & 0.283 [0.208, 0.372]
  & 0.025 \\
MV $n=1$
  & 0.760 [0.687, 0.820]
  & 0.114 [0.050, 0.240]
  & 0.290 [0.214, 0.379]
  & 0.019 \\
MV $n=3$
  & 0.760 [0.687, 0.820]
  & 0.114 [0.050, 0.240]
  & 0.290 [0.214, 0.379]
  & 0.019 \\
MV $n=5$
  & 0.753 [0.681, 0.814]
  & 0.159 [0.079, 0.294]
  & 0.281 [0.206, 0.369]
  & 0.019 \\
$B=10$
  & ---  & ---  & ---
  & 1.000 [0.977, 1.000] \\
$B=50$
  & ---  & ---  & ---
  & 1.000 [0.977, 1.000] \\
$B=75$
  & 0.753 [0.680, 0.815]
  & 0.100 [0.040, 0.231]
  & 0.298 [0.222, 0.388]
  & 0.044 \\
\textbf{$B=100$}
  & \textbf{0.768 [0.695, 0.827]}
  & \textbf{0.095 [0.037, 0.221]}
  & \textbf{0.283 [0.208, 0.372]}
  & 0.037 \\
\textit{$B=124$}
  & \textit{0.761 [0.688, 0.822]}
  & \textit{0.116 [0.051, 0.245]}
  & \textit{0.286 [0.210, 0.375]}
  & \textit{0.037} \\
\textit{$B=150$}
  & \textit{0.761 [0.688, 0.822]}
  & \textit{0.116 [0.051, 0.245]}
  & \textit{0.286 [0.210, 0.375]}
  & \textit{0.037} \\
  \multicolumn{5}{c}{\textit{$\star$ higher due to mislabeled examples -- see Discussion}} \\
\midrule
\multicolumn{5}{c}{\textbf{SWMH (N=250)}} \\
\midrule
\textbf{Condition} & \textbf{Accuracy} $\uparrow$ & \textbf{FPR} $\downarrow$& 
\textbf{SW FNR} $\downarrow$ & \textbf{Esc.} $\downarrow$ \\
\midrule
Single-agent
  & 0.790 [0.735, 0.836]
  & 0.237 [0.184, 0.301]
  & 0.100 [0.044, 0.214]
  & 0.008 \\
MV $n=1$
  & 0.790 [0.735, 0.836]
  & 0.237 [0.184, 0.301]
  & 0.100 [0.044, 0.214]
  & 0.008 \\
\textit{MV $n=3$}
  & \textit{0.795 [0.741, 0.841]}
  & \textit{0.231 [0.178, 0.295]}
  & \textit{0.100 [0.044, 0.214]}
  & \textit{0.004} \\
\textit{MV $n=5$}
  & \textit{0.791 [0.736, 0.837]}
  & \textit{0.236 [0.183, 0.300]}
  & \textit{0.100 [0.044, 0.214]}
  & \textit{0.004} \\
$B=10$
  & ---  & ---  & ---
  & 1.000 [0.985, 1.000] \\
$B=50$
  & ---  & ---  & ---
  & 1.000 [0.985, 1.000] \\
\textit{$B=75$}
  & \textit{0.799 [0.745, 0.845]}
  & \textit{0.227 [0.174, 0.291]}
  & \textit{0.100 [0.044, 0.214]}
  & \textit{0.024} \\
\textbf{$B=100$}
  & \textbf{0.798 [0.744, 0.844]}
  & \textbf{0.227 [0.174, 0.291]}
  & 0.100 [0.044, 0.214]
  & 0.008 \\
\textit{$B=124$}
  & \textit{0.794 [0.740, 0.840]}
  & \textit{0.232 [0.179, 0.296]}
  & \textit{0.100 [0.044, 0.214]}
  & \textit{0.008} \\
\textit{$B=150$}
  & \textit{0.795 [0.741, 0.841]}
  & \textit{0.231 [0.178, 0.295]}
  & \textit{0.100 [0.044, 0.214]}
  & \textit{0.004} \\
\bottomrule
\end{tabular}
\caption{Results on AEGIS 2.0 (N=161, top) and SWMH (N=250, 
bottom) with 95\% Wilson CIs. \textbf{Bold} indicates best 
result. \textit{Italics} indicate conditions whose FPR 
confidence intervals overlap with $B=100$. $B=10$ and $B=50$ 
escalate all inputs and produce no classified outputs.}

\label{tab:results_combined}
\end{table}

\subsection{Experiment 1: AEGIS 2.0}

Table~\ref{tab:results_combined} reports accuracy, false positive 
rate, false negative rate, and escalation rate across all ten 
conditions on the AEGIS 2.0 behavioral health subset (N=161), 
with 95\% Wilson score confidence intervals throughout.

\paragraph{Accuracy.}
Adaptive sampling achieves the highest accuracy, though differences 
across valid conditions are small and confidence intervals overlap 
substantially. $B=100$ reaches 0.768 (95\% CI [0.695, 0.827]), 
compared to 0.752 for single-agent and 0.760 for MV $n=1$ and 
MV $n=3$. Accuracy alone does not differentiate conditions on 
this dataset; the primary differentiator is false positive rate.

\paragraph{False positive rate.}
Adaptive sampling produces the largest reduction in false positives 
of any condition evaluated. $B=100$ achieves FPR = 0.095 
(95\% CI [0.037, 0.221]), compared to 0.159 
(95\% CI [0.079, 0.294]) for single-agent, a 40\% reduction in 
incorrect flagging of safe content. All multi-agent conditions 
reduce FPR relative to single-agent, with $B=100$ achieving the 
largest reduction. In behavioral health settings, incorrectly 
flagging safe content increases clinician burden and risks alert 
fatigue, which can reduce vigilance toward genuinely high-risk cases.

\paragraph{False negative rate.}
FNR does not improve with additional sampling and is flat across 
all valid conditions. FNR stays between 0.281 and 0.290 regardless 
of condition, suggesting the ceiling reflects label quality rather 
than sampling strategy; we examine the specific misclassified 
examples in the \hyperlink{discussion}{Discussion}.

\begin{figure}[h]
\centering
\includegraphics[width=0.6\columnwidth]{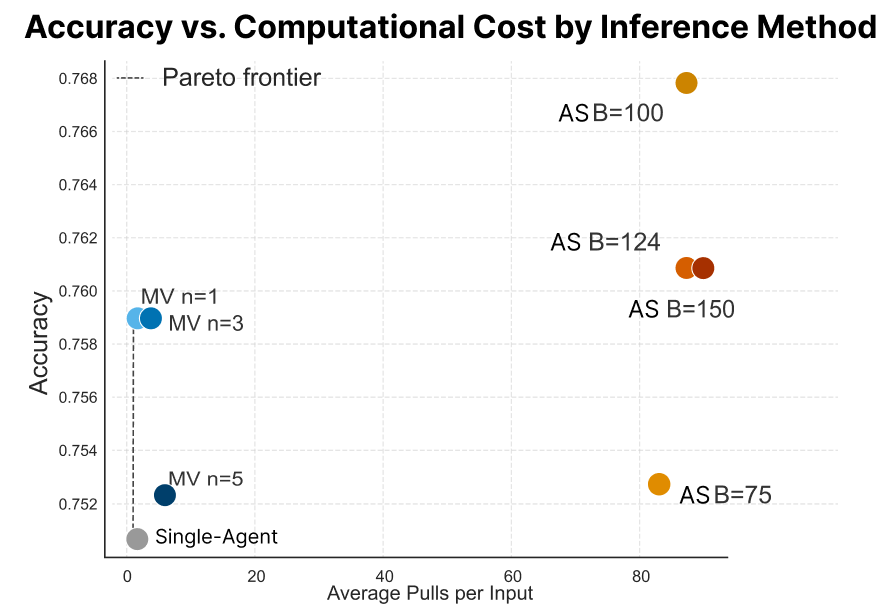}
\caption{Accuracy versus average pulls per input on AEGIS 2.0. 
Points to the left use fewer LLM calls per input; points higher 
achieve better accuracy. MV $n=1$ and MV $n=3$ lie on the Pareto 
frontier, offering the best accuracy per unit compute. Adaptive 
sampling conditions cluster at 80--90 average pulls, achieving 
higher accuracy than single-agent at substantially greater cost. 
The gap between the majority vote and adaptive sampling clusters 
reflects the cost of the reliability guarantee: spending more 
calls per input is the tradeoff the 40\% false positive reduction 
shown in Table~\ref{tab:results_combined}.}
\label{fig:efficiency}
\end{figure}

\begin{figure}[h]
\centering
\includegraphics[width=0.65\columnwidth]{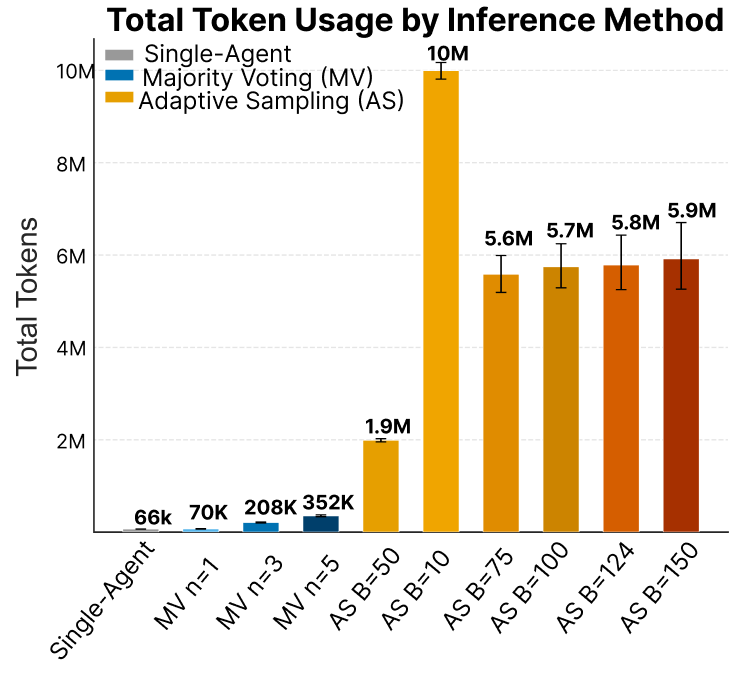}
\caption{Total token usage by condition on AEGIS 2.0. The compute 
cost of adaptive sampling reflects a deliberate trade-off: more 
calls per input is what enables the 40\% false positive reduction. 
$B=50$ exposes a gap in the current architecture: a node that 
cannot converge still consumes its full budget across all three 
pipeline stages, spending 9.9M tokens while producing no 
classified outputs, more than any other condition including 
$B=100$.}
\label{fig:tokens}
\end{figure}

\paragraph{Budget sweep.}
Performance stabilizes at $B=100$, with smaller budgets producing 
either no valid outputs or suboptimal false positive rates. $B=10$ 
and $B=50$ escalate all inputs; with $K=3$ active arms, a budget 
of 50 yields approximately 16 pulls per arm, too few for 
elimination to fire. $B=75$ is the first budget producing valid 
classifications at FPR = 0.100 (95\% CI [0.040, 0.231]), 
improving over single-agent but falling short of $B=100$. 
$B=124$ and $B=150$ offer no further improvement despite 
additional compute. The inflection between $B=75$ and $B=100$ 
provides empirical grounding for the sample complexity bound in 
\hyperlink{app:dkw}{Appendix~A}: inputs in this dataset require 
roughly 80 to 90 total pulls per arm for reliable elimination. Figure~\ref{fig:efficiency} shows the accuracy-versus-compute 
Pareto scatter, where $B=100$ achieves the highest accuracy at 
substantially higher compute cost than majority vote. 
Figure~\ref{fig:tokens} shows total token usage by condition.

\subsection{Experiment 2: SWMH Generalization}

Table~\ref{tab:results_combined} reports results on the stratified 
SWMH sample (N=250), with SuicideWatch FNR as the primary 
clinical metric.

\paragraph{Accuracy.}
Accuracy is broadly consistent across valid conditions and does 
not differentiate them on this dataset. Values range from 0.790 
for single-agent and MV $n=1$ to 0.799 for $B=75$, with 
overlapping confidence intervals throughout. The primary 
differentiator is FPR, where $B=100$ again achieves the lowest 
value.

\paragraph{SuicideWatch detection.}
The pipeline correctly identifies 90\% of SuicideWatch posts 
across all valid conditions, with the same 5 posts missed 
consistently. All valid conditions achieve SW FNR = 0.100 
(95\% CI [0.044, 0.214]). The consistency of the missed posts 
across all conditions suggests these examples are genuinely 
ambiguous rather than a failure of any particular sampling 
strategy; we examine them in \hyperlink{discussion}{discussion}.

\paragraph{False positive rate.}
Adaptive sampling again achieves the lowest FPR, consistent 
with the AEGIS 2.0 finding. FPR is substantially higher on 
SWMH than AEGIS 2.0 across all conditions, ranging from 0.227 
to 0.237, reflecting label noise in the safe subreddits: posts 
from depression, anxiety, and bipolar communities contain 
distressing content that the C-SSRS grounded prompt flags as 
potentially unsafe even when subreddit-level labels classify 
them as safe. $B=100$ achieves the lowest FPR at 0.227.

\paragraph{Budget sweep.}
Earlier arm elimination on SWMH reflects clearer input signals 
under the C-SSRS prompt. Budget conditions converge at 77--78 
average pulls per input on SWMH, compared to 86--89 on AEGIS 
2.0, suggesting SWMH posts are more clearly safe or unsafe on 
average. The budget sweep shows no measurable benefit beyond 
$B=100$, consistent with the AEGIS 2.0 finding.

\section{Discussion}
\hypertarget{discussion}{}

Fixed-sample majority voting applies the same number of LLM calls to 
every input regardless of difficulty. For behavioral health evaluation, 
where errors are asymmetric and inputs vary widely in ambiguity, this 
is a practical limitation. The adaptive sampling approach evaluated here 
allocates effort based on input difficulty and escalates when no 
confident decision is reached within budget. On AEGIS 2.0, this reduces 
false positives by 40\% relative to single-agent without affecting 
false negative rates.

In deployed settings, false positives translate directly to unnecessary 
escalations. At scale, high escalation rates contribute to alert 
fatigue, a known problem in clinical decision support that can reduce 
clinician responsiveness to genuine risk~\citep{ancker2017effects}. The 
reduction observed here is modest in absolute terms but consistent 
across both datasets evaluated.

The false negative rate does not improve with additional sampling. On 
AEGIS 2.0, FNR stays between 0.283 and 0.290 across all conditions, 
and on SWMH it holds at 0.100 regardless of budget. The same inputs 
are missed consistently, which suggests the ceiling is driven by 
annotation quality rather than sampling strategy. Inspection of the consistently misclassified examples on AEGIS 2.0 
reveals annotation noise: several posts are appropriate chatbot 
refusals that were labeled unsafe, and others contain content 
outside the scope of C-SSRS criteria entirely. These same examples 
are missed by every condition from single-agent to AS $B=150$, 
confirming the ceiling is a property of the labels rather than the 
sampling strategy. The subreddit-level labels on SWMH introduce 
similar noise. Improving ground truth annotation in behavioral 
health AI benchmarks remains an open problem.

$B=50$ uses more total tokens than any other condition, including 
$B=100$, despite producing no classified outputs. With three active 
arms and a budget of 50, each node makes roughly 16 pulls per arm 
before exhausting its budget, failing to eliminate any arm, and 
passing the input to the next node. Risk and Legal each repeat this 
process, resulting in up to 150 total pulls per input across the 
pipeline with nothing to show for it. This is a direct consequence 
of the current routing behavior: a node that cannot converge still 
consumes its full budget before escalating. Stopping early when 
budget is exhausted would avoid this entirely and is a practical 
improvement for deployment.

The FNR of 0.100 on SuicideWatch posts reflects 5 out of 50 posts 
incorrectly labeled as safe across all conditions. Reading these 
posts reveals why. Two posts explicitly deny suicidal intent despite 
expressing significant distress, so a prompt grounded in C-SSRS 
criteria correctly classifies them as safe by its own definition. 
One post is written by a caregiver asking how to help someone else, 
not a first-person account of crisis. The remaining two express 
distress indirectly without meeting explicit C-SSRS thresholds. 
These cases suggest the FNR floor reflects a mismatch between the 
explicit criteria encoded in the prompt and the broader range of 
clinical signals a clinician would recognize in context.

\section{Limitations and Future Work}

\textbf{Routing behavior on budget exhaustion.} The current system
continues routing through subsequent nodes even when a node exhausts
its sampling budget without convergence. A Worker node that cannot
confidently label an input after $B$ pulls passes the input to Risk,
which may also exhaust its budget, which then passes to Legal,
resulting in up to $3B$ LLM calls for an input that ultimately reaches
human review regardless. This is content-wise conservative as more
information is gathered before escalation, but computationally
expensive. A natural extension is early termination on budget
exhaustion: if a node cannot converge within budget, skip subsequent
nodes and escalate immediately to human review. This would reduce
compute cost on genuinely ambiguous inputs at the cost of reduced
multi-node signal. Separately, the current system does not distinguish
between budget exhaustion, where the algorithm is uncertain, and an
explicit \texttt{escalate} label, where the model is confident the
input requires specialist review. Tracking which kind of escalation
occurred at each node would enable more nuanced routing decisions and
is a direction for future work.

\textbf{Compute cost of adaptive sampling.} Budget conditions use
approximately 80--90 average pulls per input compared to 1--5 for
majority vote, representing a 20--30$\times$ increase in API calls and
associated cost. For organizations with limited compute budgets,
majority vote $n=3$ offers comparable accuracy at a fraction of the
cost. The FPR benefit of adaptive sampling must be weighed against this
operational cost in deployment planning.

Several directions follow from this work. The minimum effective budget identified here, $B=100$ pulls per node, 
was consistent across both datasets evaluated. Whether this threshold 
shifts with different underlying models, temperatures, or clinical 
populations with different input difficulty distributions remains an 
open question for future work. A related engineering improvement is early termination 
on budget exhaustion: currently, a node that cannot reach a decision 
still passes the input downstream, where subsequent nodes repeat the 
same process at full cost. Stopping early and escalating directly 
would reduce compute without changing clinical behavior. Separately, 
the pipeline currently treats all escalations the same way, whether 
the algorithm ran out of budget or the model actively chose to 
escalate. Distinguishing these two cases would give clinicians more 
information about why a case was flagged for review. Finally, SWMH relies on subreddit membership as a proxy for clinical ground 
truth rather than direct annotation. Building behavioral health 
benchmarks grounded in validated instruments like C-SSRS, with labels 
assigned by clinicians, would enable more meaningful evaluation of 
false negative rates in particular, where the current ceiling may 
reflect labeling inconsistencies as much as system limitations.

\section{Conclusion}
\label{sec:conclusion}

Behavioral health AI systems increasingly operate in clinical settings 
where errors have direct consequences for patient safety. This work 
shows that replacing fixed-sample majority voting with adaptive 
sampling reduces false positives by 40\% on a self-harm screening 
benchmark, without affecting the rate of missed detections. The same 
pattern holds on a second dataset drawn from a different source, 
suggesting the finding is not specific to one benchmark. The pipeline 
also produces a formal guarantee: when it cannot reach a confident 
decision within budget, it escalates to human review rather than 
guessing. On both datasets, the false negative rate does not improve 
with more sampling, which points to label quality in current 
benchmarks as the limiting factor rather than the algorithm itself. In safety-critical clinical settings, knowing when not to decide 
is as important as knowing how to decide.


\bibliography{sample}

\newpage
\appendix

\section*{Appendix A: DKW Inequality and Node-Level Bound}
\hypertarget{app:dkw}{}

The Dvoretzky-Kiefer-Wolfowitz (DKW) inequality provides simultaneous 
error bounds across all categories of a categorical distribution without 
the $\ln K$ penalty that arises from applying Hoeffding's inequality to 
each category separately and taking a union bound.

\paragraph{Theorem (Node-Level Categorical Bound).}
Under i.i.d.\ sampling at node $s$, for any $\delta \in (0,1)$, with 
probability at least $1-\delta$ simultaneously over all $c \in \mathcal{A}$:
\[
|\hat{p}_s(c\,;\,x) - p_s(c\,;\,x)|
\leq \epsilon_n(\delta) := \sqrt{\frac{\ln(2/\delta)}{2n}}.
\]

\paragraph{Proof.}
Impose a canonical ordering $\mathcal{A} = \{c_1, c_2, c_3\}$ and encode 
each sample as $Z_k = \phi(a^{(k)}) \in \{1, 2, 3\}$. Each category 
probability is a difference of CDF values: $p_s(c_j) = F(j) - F(j-1)$, 
where $F$ is the CDF of $Z$. Therefore:
\[
|\hat{p}_s(c_j) - p_s(c_j)| 
\leq |\hat{F}_n(j) - F(j)| + |\hat{F}_n(j-1) - F(j-1)|
\leq 2\sup_t |\hat{F}_n(t) - F(t)|.
\]
Applying the DKW inequality~\citep{dvoretzky1956asymptotic, 
massart1990tight}:
\[
\Prob\!\left(\sup_t |\hat{F}_n(t) - F(t)| > \epsilon\right) 
\leq 2e^{-2n\epsilon^2}.
\]
Setting $2e^{-2n\epsilon^2} = \delta$ and solving gives 
$\epsilon = \sqrt{\ln(2/\delta)/(2n)} = \epsilon_n(\delta)$, which 
holds simultaneously over all $c \in \mathcal{A}$ with no $K$ factor. 
The union-bounded Hoeffding approach would require 
$n \geq \frac{\ln(2/\delta) + \ln K}{2\epsilon^2}$ samples; DKW requires 
only $\frac{\ln(2/\delta)}{2\epsilon^2}$, saving $\frac{\ln K}{2\epsilon^2}$ 
samples per node. For $K=3$, $\epsilon=0.05$: approximately 220 fewer 
samples per node.

\paragraph{Corollary (Correct Routing Guarantee).}
\label{cor:routing}
Under the unique best action and i.i.d.\ sampling assumptions, if 
$n \geq n^*(\delta, \Delta_s(x)) := \frac{2\ln(2/\delta)}{\Delta_s(x)^2}$, 
then with probability $\geq 1-\delta$, the empirical argmax equals the 
true best action: $\hat{c}^* = c^*(s,x)$.

\paragraph{Proof.}
By the node-level bound, with probability $\geq 1-\delta$, 
$|\hat{p}_s(c) - p_s(c)| \leq \epsilon_n(\delta)$ for all $c$. For any 
$c \neq c^*$:
\[
\hat{p}_s(c^*) - \hat{p}_s(c) 
\geq [p_s(c^*) - \epsilon_n] - [p_s(c) + \epsilon_n]
= \Delta_s(x) - 2\epsilon_n(\delta).
\]
This is strictly positive when $\epsilon_n(\delta) < \Delta_s(x)/2$, 
i.e.\ when $n \geq 2\ln(2/\delta)/\Delta_s(x)^2 = n^*$.

\paragraph{Remark.}
The minimum sample count $n^* = 2\ln(2/\delta)/\Delta_s(x)^2$ depends 
on input $x$ through the probability gap $\Delta_s(x)$. Easy inputs with 
large gaps need few samples; ambiguous inputs with small gaps require 
many. Fixed majority voting ignores this dependence entirely, applying 
the same $n$ regardless of input difficulty.

\section*{Appendix B: Good Event Proof}
\hypertarget{app:good_event}{}

\paragraph{Lemma (Good Event).}
Define the good event as the event that all confidence intervals contain 
the true probability at every round simultaneously:
\[
\mathcal{G} = \left\{\forall c \in \mathcal{A},\ \forall m \geq 1 :
|\hat{p}_s^{(m)}(c) - p_s(c)| \leq 
\sqrt{\frac{\ln(4KT_c^2/\delta)}{2m}}\right\}.
\]
Then $\Prob(\mathcal{G}) \geq 1-\delta$.

\paragraph{Proof.}
Fix one arm $c$ and one round where it has been pulled $m$ times. By 
Hoeffding's inequality:
\[
\Prob\!\left(|\hat{p}_s^{(m)}(c) - p_s(c)| > 
\sqrt{\frac{\ln(4KT_c^2/\delta)}{2m}}\right) 
\leq 2\exp\!\left(-\ln\!\left(\frac{4KT_c^2}{\delta}\right)\right) 
= \frac{\delta}{2KT_c^2}.
\]
Taking a union bound over $K$ arms and summing over all possible rounds 
$m \in \{1, 2, \ldots\}$ using the standard series bound 
$\sum_{m=1}^{\infty} \frac{1}{m^2} = \frac{\pi^2}{6} < 2$:
\[
\Prob(\mathcal{G}^c) 
\leq \sum_{c=1}^{K} \sum_{m=1}^{\infty} \frac{\delta}{2KT_c^2}
\leq K \cdot \frac{\delta}{2K} \cdot \frac{\pi^2}{6} 
< \delta.
\]
Therefore $\Prob(\mathcal{G}) \geq 1 - \delta$. 

\paragraph{Proposition (Correctness of Algorithm~\ref{alg:se}).}
Under the unique best action assumption, with probability $\geq 1-\delta$, 
Algorithm~\ref{alg:se} never eliminates $c^*$. It either returns $c^*$ 
or returns \texttt{escalate}.

\paragraph{Proof.}
Under $\mathcal{G}$, arm $c^*$ is eliminated only if 
$\hat{p}_s(c) - w_c > \hat{p}_s(c^*) + w_{c^*}$ for some $c \neq c^*$. 
Under $\mathcal{G}$, this would require $p_s(c) > p_s(c^*)$, 
contradicting the assumption that $c^*$ is the unique best arm. 
Therefore $c^*$ is never eliminated. If the budget is exhausted before 
a single arm survives, the algorithm returns \texttt{escalate} rather 
than forcing a potentially wrong label.

\section*{Appendix C: MDP Formulation and Regret Proofs}
\hypertarget{app:mdp}{}

\paragraph{MDP Formulation.}
We model the evaluation workflow as a bounded-horizon MDP 
$\mathcal{M} = (\mathcal{S}, \mathcal{A}, P, R, \tau_{\max})$ structured 
by the DAG $G$. The state space $\mathcal{S}$ corresponds to the three 
agent nodes plus terminal states 
$\mathcal{S}_{\mathrm{term}} = \{\texttt{safe}, \texttt{unsafe}, 
\texttt{human-review}\}$. Transitions $P(s' \mid s, a)$ are restricted 
to DAG edges: terminal labels route to the corresponding terminal state, 
while \texttt{escalate} routes to the next agent. The reward function 
$R(s, a)$ captures evaluation quality at each step with 
$|R(s,a)| \leq R_{\max}$, and episodes have maximum length 
$\tau_{\max} \leq 3$. Cumulative regret over $T$ episodes is:
\[
\mathrm{Reg}(T) = \sum_{t=1}^T 
\bigl[V^{\pi^*}(s_0, x_t) - V^{\pi}(s_0, x_t)\bigr],
\]
where $\pi^*$ is the oracle policy knowing all true label distributions.

\paragraph{Theorem (Majority-Vote Regret).}
\label{thm:mv-regret}
Under fixed majority voting with $n$ samples per node:
\[
\mathrm{Reg}_{\mathrm{MV}}(T)
\leq 2\tau_{\max} R_{\max} |\mathcal{S}| \cdot T \cdot 
e^{-n\Delta_{\min}^2/2}.
\]
This grows as $\Theta(T)$: the error probability is constant across all 
episodes because the policy never adapts.

\paragraph{Proof.}
By the node-level bound with $\epsilon = \Delta_s(x)/2$, the routing 
error probability at node $s$ is at most $2e^{-n\Delta_{\min}^2/2}$. 
Each error contributes at most $\tau_{\max} R_{\max}$ to regret. A union 
bound over $|\mathcal{S}|$ nodes and summing over $T$ episodes gives the 
stated bound. Achieving sublinear regret requires 
$n = \Omega(\ln T / \Delta_{\min}^2)$, which requires knowing 
$\Delta_{\min}$ in advance. 

\paragraph{Theorem ( Adaptive Regret).}
\label{thm:ucb-regret}
The adaptive policy $\pi_{\mathrm{UCB}}$ using Algorithm~\ref{alg:se} 
achieves:
\[
\mathrm{Reg}_{\mathrm{UCB}}(T)
\leq C \cdot \frac{\tau_{\max} R_{\max} \cdot K|\mathcal{S}| \ln T}
{\Delta_{\min}^2},
\]
for a universal constant $C > 0$, without knowledge of $\Delta_{\min}$. 
This grows as $O(\log T)$: doubling the number of patient interactions 
increases total mistakes by only $\ln 2 \approx 0.69$.

\paragraph{Proof.}
Set $\delta = 1/T$. By the sample complexity result, each node identifies 
$c^*$ after $O(K\ln(KT)/\Delta_{\min}^2)$ episodes (learning phase). 
Each learning-phase episode contributes at most $\tau_{\max} R_{\max}$ 
to regret; exploitation-phase episodes contribute $O(R_{\max}/T)$ each. 
Summing over $|\mathcal{S}|$ nodes gives the $O(\log T)$ bound.

\paragraph{Theorem (System-Level Regret).}
\label{thm:system}
Under the adaptive policy:
\[
\mathrm{Reg}_{\mathrm{UCB,sys}}(T)
\leq C \cdot \frac{K \cdot |\mathcal{S}| \cdot \tau_{\max} \cdot 
R_{\max} \cdot \ln T}{\Delta_{\min}^2}.
\]
The ratio of adaptive to fixed-budget regret satisfies:
\[
\frac{\mathrm{Reg}_{\mathrm{UCB}}(T)}{\mathrm{Reg}_{\mathrm{MV}}(T)}
= O\!\left(\frac{\ln T}{T \cdot e^{-n\Delta_{\min}^2/2}}\right)
\xrightarrow{T\to\infty} 0.
\]
The adaptive policy is asymptotically dominant.

\paragraph{Proof.}
At most $\tau_{\max}$ nodes are visited per episode, and $|\mathcal{S}|$ 
nodes contribute independently to learning-phase regret, giving the 
$K|\mathcal{S}|\tau_{\max}$ factor. The ratio follows from 
$\mathrm{Reg}_{\mathrm{UCB}}(T) = O(K|\mathcal{S}|\tau_{\max}\ln T / 
\Delta_{\min}^2)$ and 
$\mathrm{Reg}_{\mathrm{MV}}(T) = \Omega(\tau_{\max} T 
e^{-n\Delta_{\min}^2/2})$.

\section*{Appendix D: Reproducibility}
\hypertarget{app:reproducibility}{}

\paragraph{Model configuration.}
All experiments use \texttt{claude-sonnet-4-6} at temperature 0.7 with 
a maximum of 10 output tokens per call. The 10-token limit is sufficient 
for the label-only output format and reduces cost substantially relative 
to unconstrained generation.

\paragraph{System prompt structure.}
Each node receives a role-specific system prompt that defines its 
clinical function (Worker: frontline screening; Risk: secondary clinical 
review; Legal: institutional compliance) and instructs the model to 
respond with exactly one of \texttt{safe}, \texttt{unsafe}, or 
\texttt{escalate}, grounded in C-SSRS criteria. The prompt includes 
definitions of each label consistent with the action space defined in 
the Methods.

\paragraph{Dataset preprocessing.}
AEGIS 2.0 examples are filtered to rows where \texttt{violated\_categories} 
contains suicide or self-harm keywords across all three splits 
(train, validation, test), deduplicated by example ID, yielding N=163 
unique examples. Two examples are excluded due to Unicode encoding errors. 
SWMH examples are sampled with \texttt{random\_state=42} using stratified 
sampling of 50 examples per subreddit.

\paragraph{Confidence intervals.}
All proportions are reported with 95\% Wilson score confidence 
intervals~\citep{wilson1927probable}, computed using the standard formula:
\[
\frac{\hat{p} + \frac{z^2}{2n} \pm z\sqrt{\frac{\hat{p}(1-\hat{p})}{n} 
+ \frac{z^2}{4n^2}}}{1 + \frac{z^2}{n}},
\]
where $z = 1.96$ for 95\% confidence and $n$ is the number of 
non-escalated examples for FPR and FNR, or total examples for 
escalation rate.

\paragraph{Code.}
Code and data preprocessing scripts will be made available upon 
acceptance.

\end{document}